\documentclass[a4paper,twoside]{article}

\usepackage{epsfig}
\usepackage{subcaption}
\usepackage{calc}
\usepackage{amssymb}
\usepackage{amstext}
\usepackage{amsmath}
\usepackage{amsthm}
\usepackage{multicol}
\usepackage{mathptmx}
\usepackage{apalike}
\usepackage{CJKutf8}
\usepackage{geometry}
\usepackage{arydshln} 
\usepackage{pslatex} 
\usepackage{enumitem} 
\usepackage{natbib, natbibspacing} 
\renewcommand{\cite}[1]{\citep{#1}} 
\usepackage{graphicx} 
\usepackage{SCITEPRESS} 

\begin{document}
\title{On The Limits To Multi-Modal Popularity Prediction on Instagram:\\ A New Robust, Efficient And Explainable Baseline}

\author{\authorname{Christoffer Riis\textsuperscript{\textsection}\sup{1}\orcidAuthor{0000-0002-4540-6691}, Damian Konrad Kowalczyk\textsuperscript{\textsection}\sup{1,2}\orcidAuthor{0000-0002-5612-0859}, Lars Kai Hansen\sup{1}\orcidAuthor{0000-0003-0442-5877}}
\affiliation{\sup{1}Technical University of Denmark, DTU Compute, Matematiktorvet 303B}
\affiliation{\sup{2}Microsoft Corporation, Business Applications Group, Kanalvej 7, 2800 Kongens Lyngby, Denmark}
\email{\{chrrii, damk, lkai\}@dtu.dk, dakowalc@microsoft.com}
}

\abstract{
%
%
Our global population contributes visual content on platforms like Instagram, attempting to express themselves and engage their audiences, at an unprecedented and increasing rate. 
In this paper, we revisit the popularity prediction on Instagram. We present a robust, efficient, and explainable baseline for population-based popularity prediction, achieving strong ranking performance. We employ the latest methods in computer vision to maximise the information extracted from the visual modality. We use transfer learning to extract visual semantics such as concepts, scenes, and objects, allowing a new level of scrutiny in an extensive, explainable ablation study. We inform feature selection towards a robust and scalable model, but also illustrate feature interactions, offering new directions for further inquiry in computational social science. Our strongest models inform a lower limit to population-based predictability of popularity on Instagram. The models are immediately applicable to social media monitoring and influencer identification.
}
\onecolumn \maketitle 
\begingroup\renewcommand\thefootnote{\textsection}
\footnotetext{These authors contributed equally}
\endgroup
\normalsize \setcounter{footnote}{0} \vfill
\section{\uppercase{Introduction}}
\noindent Social media platforms are full of societal metrics. The reach of social media postings and the mechanisms determining popularity are of increasing interest for scholars of diverse disciplines. In sociology, it can be used to understand the connection between popularity and self-esteem \cite{Social2017}; in marketing and branding, it can clarify how to best engage and communicate with customers \cite{Overgoor2017}; 
in journalism, it can be used to decide which posts to share on social media \cite{News2019a}; 
and in political science, it can be used to understand how personalised content affect popularity \cite{Political2019}. 
From a data science point of view, giving a lower bound on the limits to the predictability of human behaviour is a challenging task. In Song et al.'s seminal work on limits to mobility prediction, they argue that there is a huge gap between population and individual prediction: while individual predictability is high, population-based predictability is much lower \cite{song2010limits}. 
Well-aligned with \citet{song2010limits}, very high popularity predictability of individuals' postings on Instagram are found by combining individualised models \cite{Gayberi2019}.
Oppositely, this paper focuses on Instagram popularity prediction as the hard problem of predicting popularity using population models. Following the generality track in the population models, we will not restrict the analysis to any specific segment. Instead we will use a general segment, which is in sharp contrast to previous studies on Instagram predicting popularity  \cite{Mazloom2016,Mazloom2018,Overgoor2017}. 
To the best of our knowledge, we are the first to use population models to predict popularity on Instagram as a regression and ranking problem with a general segment. 
In this paper, we further investigate and explain the visual modality and its potential for popularity ranking. Our contributions can be summarized as follows:
\begin{enumerate}[noitemsep]
    \item we advance user-generated visual modality representation with a novel and rich set of features, and provide detailed explanations of their impact,
    \item we provide two new popularity models for Instagram, which achieve strong ranking performance in a robust and explainable way, and finally
    \item we offer a new lower bound to predictability of Instagram popularity with the above general population models.
\end{enumerate}
Additionally, our modelling contributions are bridging previous studies of the visual modality on Instagram \cite{Mazloom2016,Mazloom2018,Gayberi2019,Overgoor2017,Rietveld2020} through a clarification of the influence of different visual aspects on popularity alongside an investigation of the role of four different feature sets in a comprehensive ablation study.

\section{\uppercase{Related work}}
\noindent With the ever increasing volume of multi-modal uploads to the social media platforms, the challenge of predicting the popularity of user-generated content inspires multi-modal approaches including content (metadata), author, textual, and visual information. 
%
Content and user information are used with a Gradient Boosting Machine (GBM) to achieve excellent results \cite{Kang2019}. In multiple ablation studies, it is reported that the content and user information indeed are the strongest predictors among the four modalities \cite{Ding2019a,He2019,Wang2018}. These studies also show how the modelling of textual content improve the performance but show mixed performance for the visual content, suggesting that care has to be exercised when combining the modalities. In the following, we pay extra attention to the visual modality and how it is modelled in earlier work.

\citet{Khosla2014} find performance gains from combining low-level features and semantic features such as objects. Moreover, they conclude that scenes, objects and faces are good as predictors for image popularity. Similarly, other studies consider both colour features, analysis of the scenery, and the number of faces in the images \cite{McParlane2014}, and visual information extracted form a pre-trained neural network \cite{Cappallo2015}. Both studies show promising results for the visual modality as a descriptor for popularity prediction.

Extant recent work considers high level visual information such as concepts, scenes, and objects derived by transfer learning in the form of neural networks trained for classification or object detection tasks \cite{Gayberi2019,Gelli2015,Mazloom2018,Ortis2019}. 
An overview is shown in Table \ref{tab:visual_features}.
\citet{Gayberi2019} suggest that objects and categories are important features in order to utilise the visual modality in the best way possible and therefore propose to use the MS COCO Model \cite{Caesar2018} for object detection. 
\citet{Gelli2015} use a pre-trained network for object detection to extract high-level features and objects. Their quantitative analysis shows how the visual features complement the strong information from the content and author features.
\citet{Mazloom2018} focus on popularity prediction within different categories such as action, animal, people, and scene. They show how human faces and animals are important for popularity prediction.
\citet{Ortis2019} hypothesise that semantic features of the images such as objects and scenes have an impact on the performance and therefore, they extract predictions from two different neural networks. 
Another approach is to use an image-captioning model to extract the high level information \cite{Hsu2019,Zhang2018}.
%
%
\begin{table}[ht]
\caption{Summary of the use of concepts, scenes, and objects extracted from the visual modality.}
\label{tab:visual_features}
\centering
\scriptsize
\begin{tabular}{lccccccl}
              & Concepts  & Scenes    & Objects    \\ \hline
\citet{Gayberi2019}   & X &           & X \\
\citet{Gelli2015}     & X &           &           \\
\citet{Khosla2014}    & X &           &          \\
\citet{Mazloom2018}   & X &           &           \\
\citet{Mazloom2016}   & X &           & X \\
\citet{McParlane2014} &           & X & X \\
\citet{Ortis2019}     & X & X &           \\
\citet{Overgoor2017}  & X &           &          \\
\citet{Rietveld2020}  &           &           & X\\ \hline
\textbf{This study} &  X &  X & X \\ \hline
\end{tabular}
\end{table}
Visual features include brightness, style, and colour. Quantifying the aesthetics of images in popularity prediction is seen in several papers \cite{Chen2019,Ding2019a,Hidayati2017,Mazloom2016}. 
\citet{Chen2019} propose to use moments 
to quantify the style and colour.
\citet{Ding2019a} use a network directly pre-trained to access the image aesthetics.
\citet{Hidayati2017} hypothesise that visual aesthetics are important information and therefore, they extract several high-level semantic features such as brightness, clarity, colour, and background simplicity. 
\citet{Mazloom2016} directly extract image aesthetics as a 42-dimensional binary vector given by the content information from Instagram in the form of the feature \textit{filter}.
Another high-level feature is visual sentiment, which can be directly assessed with neural networks \cite{Gelli2015,Mazloom2016}. However, we hypothesise that these features are captured in the high-level features from a deep neural network and consequently, we do not apply this approach. 

\noindent In multiple works, visual features are extracted implicitly by neural network embeddings pre-trained for general object recognition tasks. Many use a deep neural network pre-trained on ImageNet \cite{ImageNet} for classification
(e.g. \cite{Mazloom2018,Ortis2019,Wang2018}). 
It is most common to use the embeddings from the last pooling layer with either 1024 or 2048 individual real-valued features, depending on the network structure \cite{Ding2019a,Mazloom2018,Mazloom2016,Overgoor2017}. 
\citet{Ortis2019} extract high-level features from three different networks by considering the last two activation layers. The three networks are pre-trained predicting classes, adjective-noun pairs, and object and scenes.
\citet{Wang2018} use features from a network pre-trained on ImageNet and afterwards fine-tune the network for popularity prediction.

While several papers deploy transfer learning to access semantic and high-level features, recent work applies end-to-end models on the visual modality \cite{Ding2019,Zhang2019}. \citet{Zhang2019} investigate the effectiveness of using neural networks in the modelling of image popularity. They hypothesise that the text features have a stronger predictive power than the visual features. With a six-layer end-to-end network, they outperform their baseline comprised of a pre-trained deep neural network 
together with Support Vector Regression and show how their network is comparable with the text-based embeddings methods. 
\citet{Ding2019} investigate the contribution of the visual content in popularity prediction by training a deep neural network to predict the intrinsic image popularity. By dividing posts into different pairs giving user statistics, upload time, and captions, they train the network with a Siamese architecture. Through a qualitative analysis and a psycho-physical experiment, they show how their intrinsic image popularity assessment model (IIPA) achieves human-level performance.
\\
\indent
\textbf{Our design space}: We aim to construct a new image feature extractor building upon recent work utilising deep learning (e.g \cite{Ding2019a,He2019,Ortis2019}). 
In recent years, the application of deep learning and neural networks have grown intensively as the field of computer vision has advantaged within classification \cite{EfficientNet} and 
object detection \cite{YOLOv3} among others. 
Accordingly, we propose to use transfer learning with the most recent networks of computer vision to represent visual information and measure its importance in predicting popularity on social media. In relation to previous use of transfer learning and embeddings
(e.g. \cite{Ding2019,Mazloom2016,Ortis2019}),
we improve the explainability of the embeddings by constructing them as the input to the classifier softmax, i.e. the last layer prior to the softmax, so each feature has a class label associated.

\begin{figure}[ht]
    \centering
    \includegraphics[width=0.97\columnwidth]{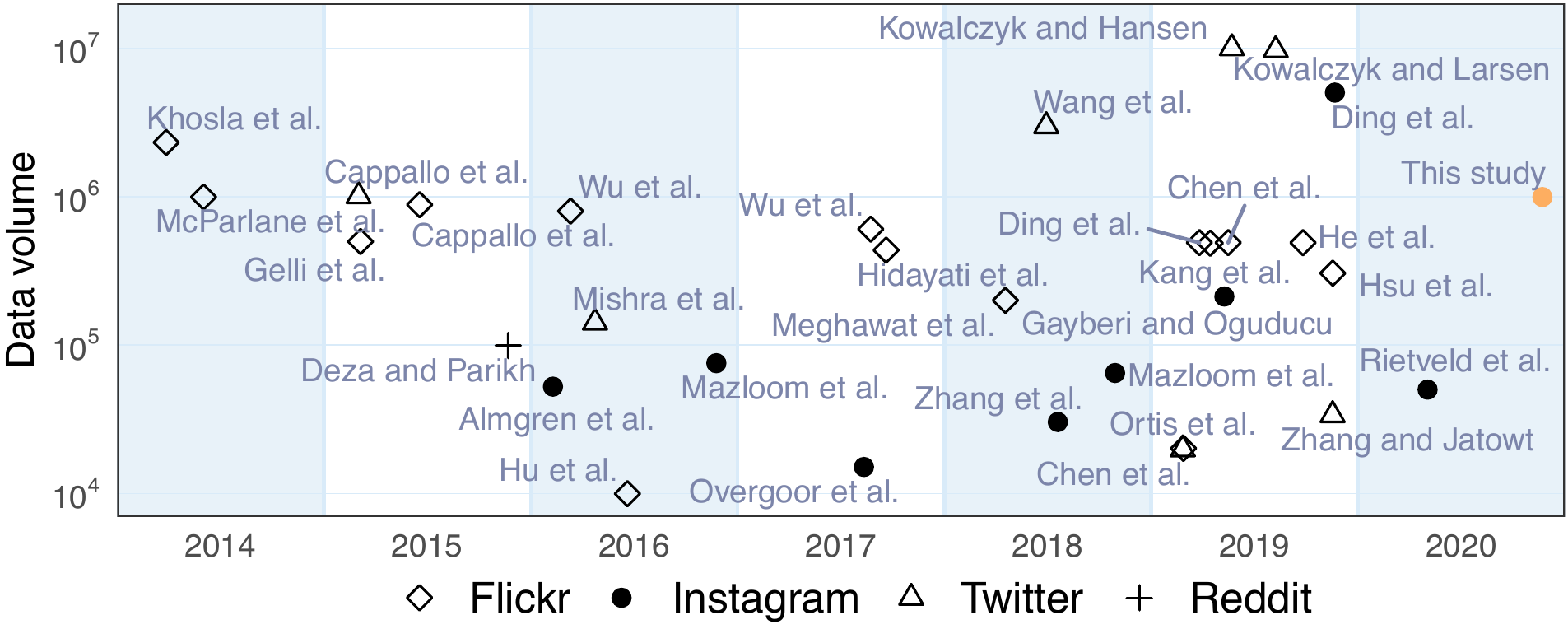}
    \caption{Different sizes of data sets have been used on the different platforms. This study (orange point) with 1 million samples is among the largest popularity prediction studies on both Instagram and social media in general. Points are shifted left or right for visual clarity.}
    \label{fig:data_sizes}
\end{figure}

Networks pre-trained for different tasks have different internal representations, which means that the high-level features will be complementary in describing images \cite{Zhou2014}. 
Therefore, we will use the deep neural network EfficientNet-B6 \cite{EfficientNet} pre-trained for classification, Places365 ResNet-18 \cite{Places365} pre-trained for scene classification, and YOLOv3 \cite{YOLOv3} pre-trained for object detection.
We adopt the model IIPA \cite{Ding2019} to assess the intrinsic image popularity directly.
Besides introducing the state-of-the-art networks EfficientNet, Places365, and YOLOv3 in popularity prediction, these pre-trained models give a novel combination (also shown in Table \ref{tab:visual_features}) of the visual semantics concepts, scenes, and objects. 
The combination of the four complementary models leads to a rich image representation, instrumental for advancing the popularity prediction on Instagram. We maximise the semantic diversity of the representation to boost the final model's ranking performance and engagement explainability simultaneously. To test the final model, we gathered one million posts from Instagram (more details in the sections on methods). Figure \ref{fig:data_sizes} shows that the size of our data set is among the largest data sets on both Instagram and other social media platforms. 
%

\noindent Finally, we define our scope of popularity prediction and measurement. 
There exist multiple ways to address popularity prediction on social media. Previous work predict the number of mentions for a specific event \cite{Chen2019a}; look at the popularity over time or as a cascade \cite{Almgren2016,Mishra2016,Ortis2019,Wu2016,Wu2017}; 
define it as a binary classification problem \cite{Deza2015,McParlane2014,Zhang2018}; but the main focus in popularity prediction on social media is to predict the number of likes, shares, views, etc., as a regression and ranking problem
(e.g. \cite{Chen2019,He2019,Kowalczyk2019}). 
In this paper, we address popularity prediction as a regression and ranking problem. As popularity measurement, we follow the majority of the literature and use the number of likes as our response variable
(e.g \cite{Ding2019,Rietveld2020,Zhang2018}). 

\section{\uppercase{Methods}}\label{sec:method}
In this section, we first describe the 1M size data set and how it was gathered. Next, we outline the feature extraction by going through the social features and the enhanced visual feature extractor. 
Then, we describe the gradient boosting machine used for prediction. Lastly, we briefly introduce our use of the explainability tool SHAP \cite{Lundberg2017}.

As mentioned by several studies, there does not exist a publicly available data set for Instagram (e.g. \cite{Gayberi2019,Overgoor2017,Zhang2018}). 
Similar to previous studies
(e.g. \cite{Gayberi2019,Mazloom2018,Rietveld2020}),
we scraped Instagram and created a multi-modal data set for this study specifically. The data set consists of one million image posts gathered from 2018-10-31 to 2018-12-11.
The data set is neither categorical nor user-specific and can thus be seen as a general subset of all image posts on Instagram. However, we are aware of the inevitable bias that lies in the discard of non-public posts. The image, engagement signal, and social information were picked up 48 hours after upload time.

Previous studies show that the performance of popularity prediction benefits from a multi-modal approach \cite{Ding2019a,Hsu2019,Wang2018}. Therefore, we extract features from several information sources. Overall, the features collected from each post can be divided into social features and visual features.
The social features are branched into author, content, and temporal features. 
Among the author features, we extract how many followers the user has, how many other users the user follows, and the number of posts the user has made. In order to stabilise the variance, we log-normalise these three variables (e.g. \cite{Ding2019a,Gayberi2019,Kowalczyk2019a}). 
The transformation of a variable $x$ is given as follows by first log transforming the variable $x_{log} = \log( x +1 )$
and then subtracting the mean
\begin{equation}\label{eq:mean_sub}
    x_{transformed} = x_{log} - mean({x}_{log}).
\end{equation}
Furthermore, we augment the features by computing the ratios \textit{follower per post} and \textit{follower per following} \cite{Kowalczyk2019}. 
Regarding the content features, we extract image filter, number of users tagged, whether the user liked the post, if geolocation is available, language, the number of tags, and the length of the caption measured in words and characters. From the language features, we augment the data with \textit{is English}.
Regarding the temporal features, we extract the feature consisting of the date and time for posting and split it into \textit{posted date}, \textit{posted week day}, and \textit{posted hour} \cite{Kowalczyk2019a}. 
\textit{User ID} and \textit{activity ID} are discarded as irrelevant for the population-based approach, effectively anonymizing the training.
%
%
In creating a comprehensive visual feature extractor, we use transfer learning and deploy four pre-trained neural networks in order to describe concepts, scenes, objects, and intrinsic image popularity.\\
\indent\textit{Concept features}: To extract concept features, we use the state-of-the-art model EfficientNet-B6 \cite{EfficientNet} pre-trained on ImageNet \cite{ImageNet}. We use the values in the last layer prior to the softmax normalization layer. This provides a 1000-dimensional vector each entry corresponding to a high level object class label.\\
\indent\textit{Scene features}: We extract a diverse set of scene features by using Places365 ResNet-18 \cite{Places365}. We use the values of the last layer prior to softmax normalization. This provides a 365-dimensional interpretable vector of scene concepts, a 102-dimensional feature vector of SUN scene attributes \cite{Patterson2012}, and a single entry indicating if the scene is indoors or outdoors.\\
\indent\textit{Object features}: YOLOv3 \cite{YOLOv3} pre-trained on COCO \cite{Lin2014} is used to detect multiple occurrences of 80 different objects. For each object, we count the number of instances providing a 80-dimensional `bag-of-objects' histogram of object occurrences.\\
\indent\textit{Intrinsic image popularity}: Here, we adopt the model IIPA \cite{Ding2019} to directly assess the intrinsic image popularity in a single variable.

\begin{figure}[ht]
    \centering
    \includegraphics[width=0.98\columnwidth]{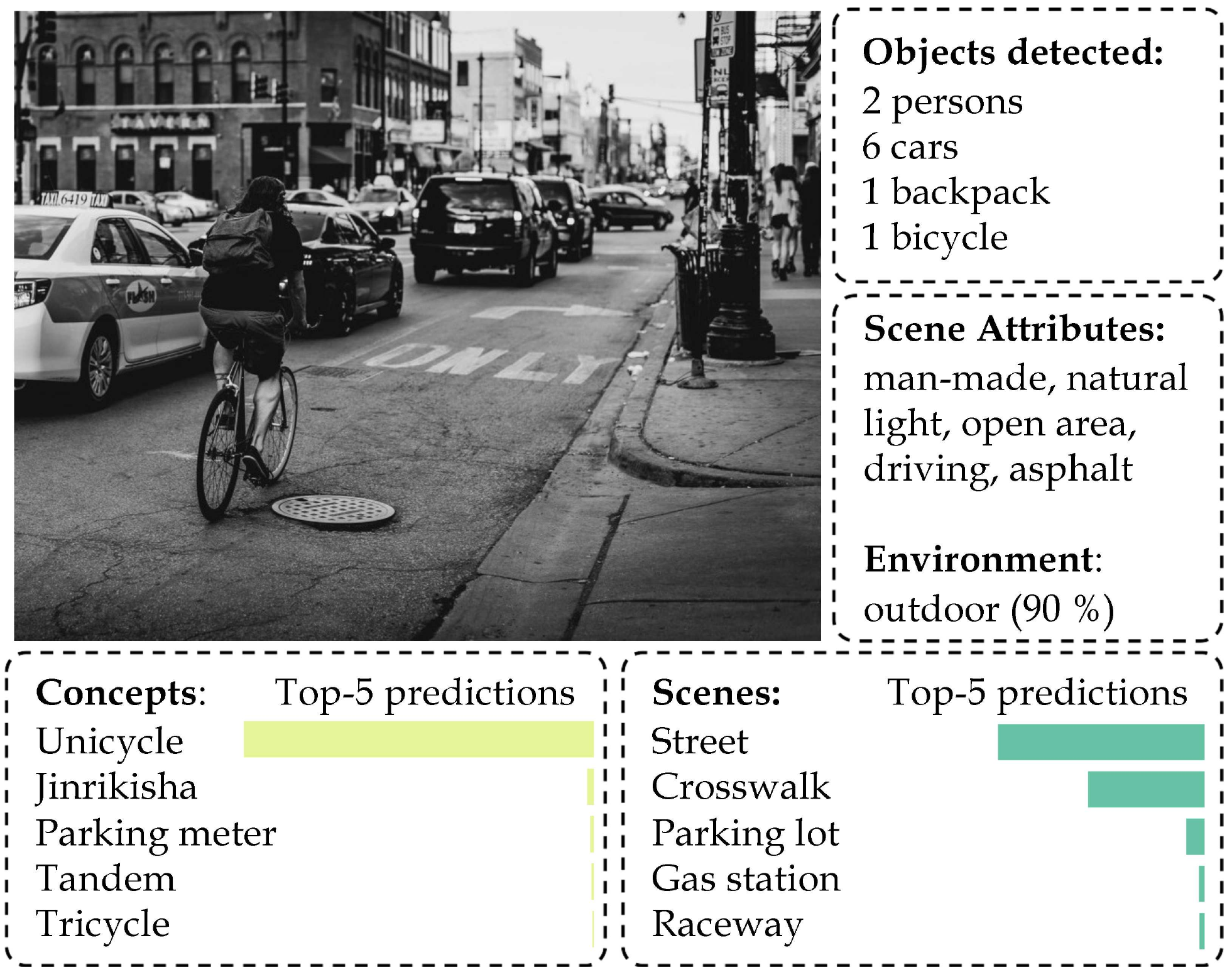}
    \caption{Example of the features extracted from an image. The associated concepts are extracted with EfficientNet, objects are detected using YOLOv3, and the associated scenes and scene attributes as well as the environment (indoor/outdoor) are extracted with Places365. Additionally, the image scores a neutral IIPA value at 1.96 on a normalised scale from -4 to 8, with a mean of 2.}
    \label{fig:predictions}
\end{figure}

In total, we have 1548 features representing concepts, scenes, objects, and the intrinsic image popularity 
resulting in an expressive and comprehensive visual feature representation. 
A feature extraction is illustrated in Figure \ref{fig:predictions}.
\\

\noindent Gradient boosting algorithms are used in social media popularity prediction
due to speed, performance and explainability (e.g. \cite{Chen2019,Gayberi2019,Kang2019}). We use the framework LightGBM \cite{LightGBM} in line with other recent studies \cite{He2019,Hsu2019,Kowalczyk2019a,Kowalczyk2019}.
LightGBM is a leaf-wise growth algorithm and uses a histogram-based algorithm to approximately find the best split. The algorithm handles integer-encoded categorical features and uses Exclusive Feature Bundling (EFB). By combining gradient-based one-side sampling and EFB, \citet{LightGBM} show how this algorithm can accelerate the training of previous GBMs by 20 times or more while achieving at par accuracy across multiple public data sets.
The number of likes is the most popular engagement signal on Instagram 
(e.g. \cite{Ding2019,Mazloom2018,Gayberi2019}). 
We choose to predict the log-normalised number of likes (transformations from eq. (\ref{eq:mean_sub})) with the Spearman's rank correlation (SRC), Root Mean Square Error (RMSE), and $R^2$ as evaluation metrics.\\

\noindent\textbf{Explainable ML}:
We use the \textit{SHAP} \cite{Lundberg2017} library to compute feature level explanations. Single Shapley value quantifies the effect on a prediction, which is attributed to a feature. Two properties of these values make them ideal for explaining our ablation study:\\
    \indent\textit{Consistency and local accuracy:} If we change the model such that a feature has a greater impact, the attribution assigned to that feature will never decrease. Features missing in the original input (i.e. removed in ablation) are attributed no importance. The values can be used to explain single predictions and to summarise the model.\\
    \indent\textit{Additivity of explanations:} Summing the effects of all feature attributions approximates the output of the original model. Additivity, therefore, enables aggregating explanations, e.g., on a group level, towards an accurate and consistent attribution for each of the modalities in the study.\\

\noindent\textbf{Model training}: We train 111 models for the ablation study (37 combinations in 3-fold cross-validation) in a distributed environment of Apache Spark. The cluster consists of 3 nodes, each powered by a 6-core Intel Xeon CPU and an NVidia Tesla V100 GPU. We perform a basic hyper-parameter tuning of LightGBM on the full combination of feature groups (denoted as YIEPACT) and fix these parameters across ablation experiments to ensure fair comparison. We cap the number of leaves at 256, set the feature sampling at every iteration to 0.5 (expecting many noisy features to slow down the training otherwise), limit the number of bins when building the histograms to 255 
and set the learning rate to 0.05. 
\section{\uppercase{Results \& main findings}}
\begin{figure}[ht]
    \centering
    \includegraphics[width=0.98\columnwidth]{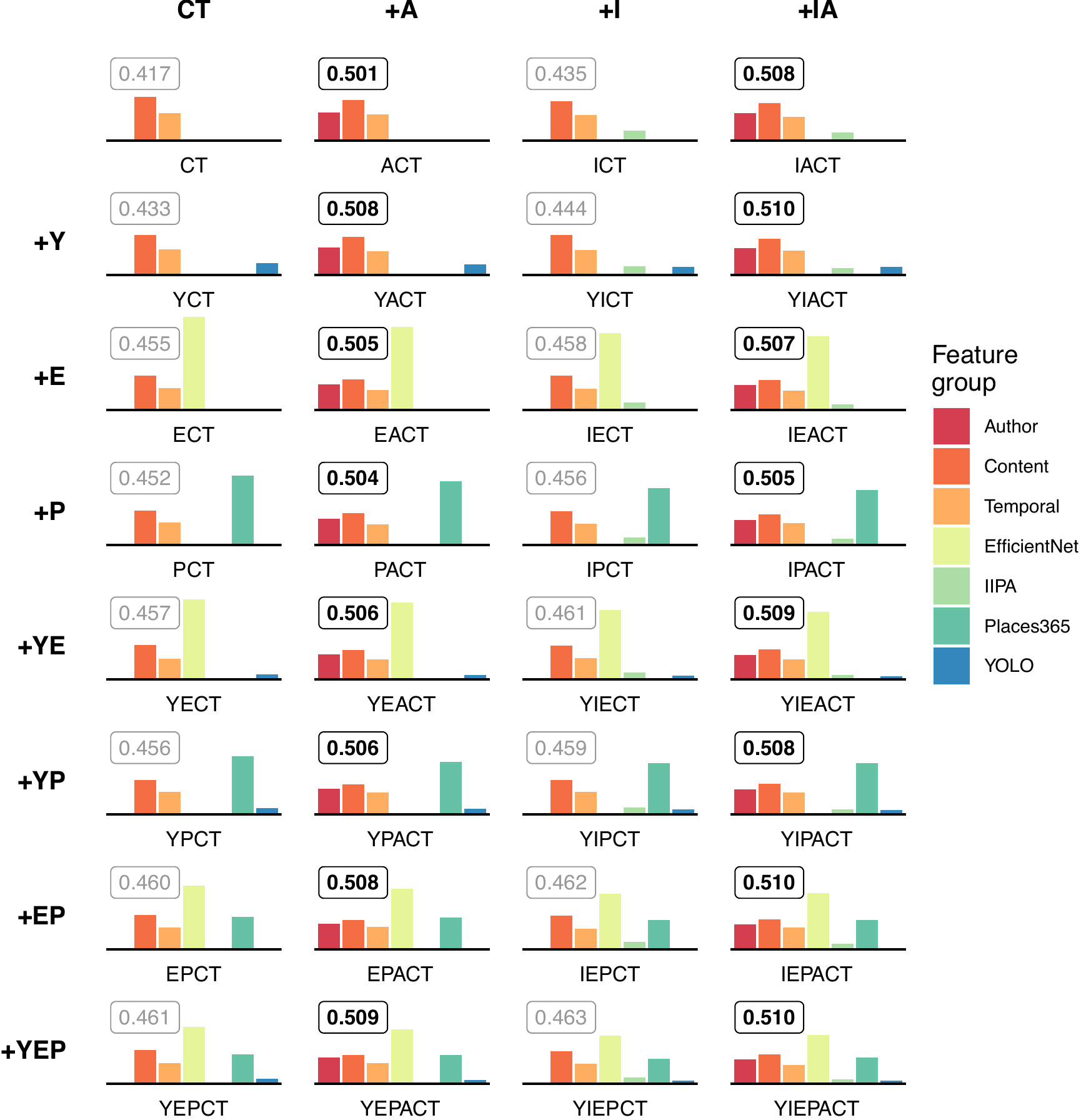}
    \caption{Average absolute SHAP value for each feature aggregated within each feature group displayed for the models. The upper left plot shows the base model with \textit{Content} (C) and \textit{Temporal} (T) features. In the columns, \textit{Author} (A) and \textit{IIPA} (I) features are added, and in the rows \textit{EfficientNet} (E), \textit{Places365} (P), and \textit{YOLOv3} (Y) - corresponding to concepts, scenes, and objects respectively - are added. The Spearman's rank correlation is shown for each model.
    }
    \label{fig:abs_shap}
\end{figure}

In Figure \ref{fig:abs_shap}, the average absolute SHAP value for each feature aggregated within each group of features are displayed for each model together with the corresponding SRC. 
%
The base model CT with content features (C) and temporal features (T) achieving an SRC of 0.417 is shown in the upper left corner. C affects the prediction more than T, since the content bar is higher than the temporal bar.\\

\noindent\textbf{Author features are essential.}
In the columns, we add author features (A), IIPA (I), and the combination of the two (IA). In the first row with the base model CT, we observe that adding I to the base model increases the performance to 0.435 SRC, whereas adding A gives a very high increase in the performance reaching an SRC at 0.501. In fact, all the rows in the second and fourth column show that these models with the author features do indeed score an SRC above 0.5. Thus, the author features appear essential for reaching strong performance.\\

\noindent\textbf{EfficientNet has the largest effect on the predictions.} In the rows below the base model CT in Figure \ref{fig:abs_shap}, the different semantic concepts (E: \textit{EfficientNet}), scenes (P: \textit{Places365}), and objects (Y: \textit{YOLOv3}) are added to the model. A comparison of the three models YCT, ECT, and PCT show that E on average, has the largest effect on the predictions. In the lower half of the column, we have the models combining these features, and again it appears that E has the largest effect. This observation can be validated across the other columns. 
%
\\

\noindent\textbf{Visual semantics are correlated.} Adding combinations of the semantic groups gives a decrease in the contribution for a single group, e.g. in YEPCT the effect of both E, P, and Y are lower than for the other models in this column. At the same time, the SRC is increased every time new features are added to the model, indicating that the different features are complementary. However, the decrease in the different bars together with the increase in the SRC also indicate that the groups are slightly correlated and that the model might learn a better representation such that some of the features within the different groups are disregarded. In other words, this illustrates the synergy between the groups and how some features are substituted by including other features. These observations can be validated across the other columns.
%
%
\\

\noindent\textbf{Object detection works better with author features.} 
In the second column in Figure \ref{fig:abs_shap}, we add A to the base model CT and observe a sudden increase in the performance reaching an SRC at 0.501.
In the first column without A, the increase in performance is higher when adding E or P instead of Y, e.g. the model EPCT achieve a higher SRC than both YECT and YPCT.
The same patterns are seen in the third column.
However, in the first column with A, the pattern is more cluttered, since YACT achieves a higher SRC than both EACT and PACT.
Moreover, adding either E or P to YACT results in a performance decrease, but adding all of them in YEPACT gives the highest performance in this column. Withal, the combination of EP in EPACT achieves the same performance as YACT. 
%
Lastly, even though both YEACT and YPACT have lower performance than YACT, adding all three visual semantics in YEPACT gives a small increase in performance. These hypotheses are validated by the fourth column. 
However, no performance gain is obtained by combing YIACT and IEPACT into YIEPACT. The three models achieve the highest observed SRC at 0.51. In sum, we see how objects together with authors features are very powerful, but also how the combination of concepts and scenes is indeed powerful with and without author features.\\
%
\begin{figure}[ht]
    \centering
    \includegraphics[width=0.98\columnwidth]{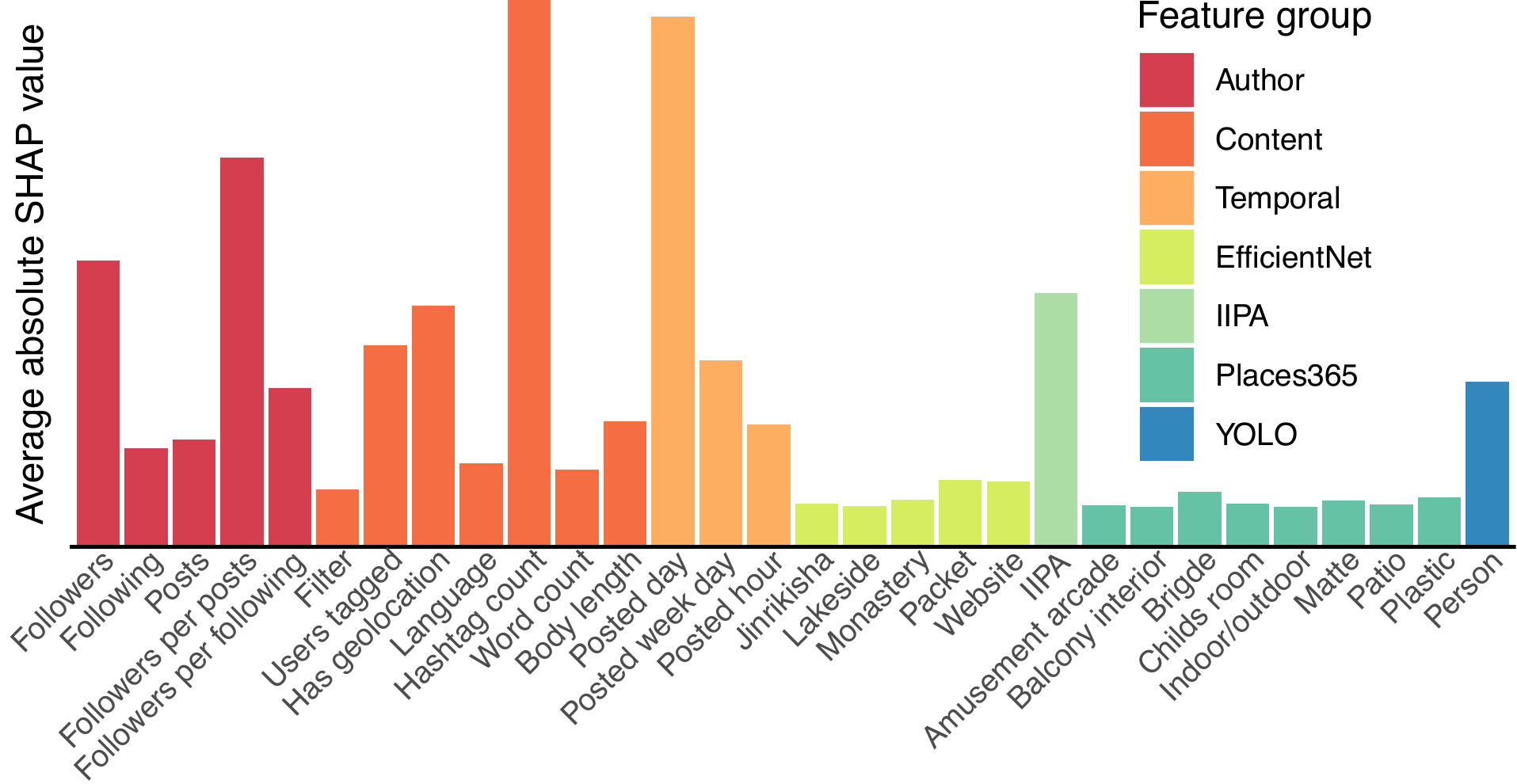}
    \caption{Average absolute SHAP value for top 30 features. The features are chosen by highest average absolute SHAP values across all models.}
    \label{fig:top_features}
\end{figure}
\newline
\noindent In the following, we will investigate the features affecting the prediction the most by finding the top-30 most prominent features based on the average absolute SHAP value across all models. More precisely, we aggregate the average absolute SHAP value for each feature across all models, and then divide by the number of times that feature is present in the models.
In Figure \ref{fig:top_features}, the top-30 features are coloured after each feature group. The features \textit{hashtag count} and \textit{posted day} have the largest average absolute SHAP value and thereby affect a prediction the most. 
The author features \textit{followers} and \textit{followers per post} come right after but more interestingly, note how the two computed ratios \textit{followers per post} and \textit{followers per following} both are high and are actually affecting the prediction more than the two features \textit{following} and \textit{posts}. The three temporal features all have a high effect on the prediction which both shows that the day of the week and the time of the day is important information for predicting the popularity. 
Among the visual features, IIPA and \textit{Person} have the largest effect and are both comparable to the social features. Yet, in general, all the visual features have a smaller effect than the social features.\\ 
%
\begin{figure}[ht]
    \centering
    \includegraphics[width=0.98\columnwidth]{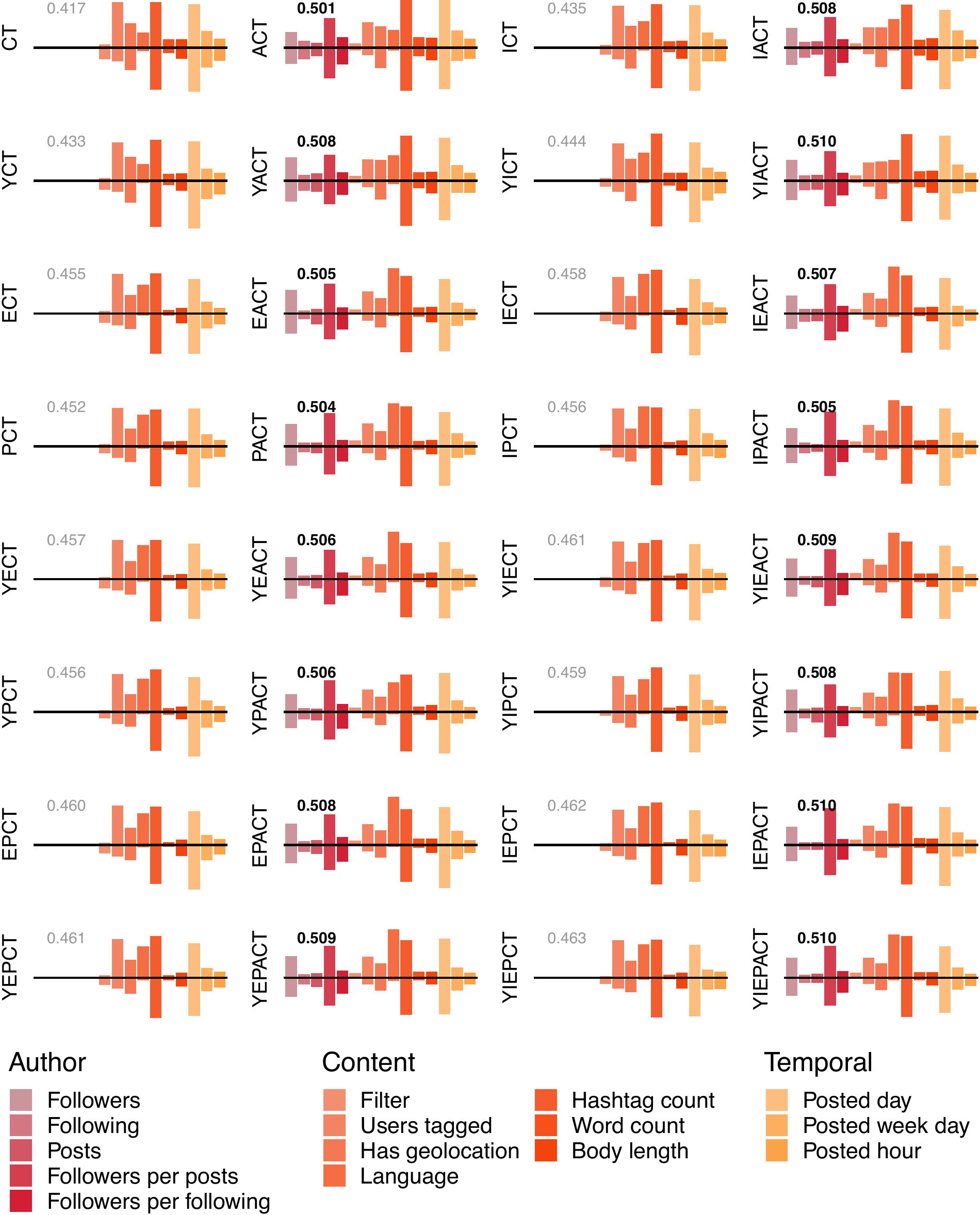}
    \caption{Average positive and negative SHAP values for most prominent social features displayed for each model.}
    \label{fig:mean_shap_social}
\end{figure}
%
%
%
\newline
The social features are explained using the SHAP values individually. We summarise the SHAP values in two numbers computed as the mean of all positive and all negative SHAP values separately. In this way, we both preserve the sign and the deviation of the SHAP values. In contrast, SHAP values of different signs will cancel out each other in a regular mean calculation. In Figure \ref{fig:mean_shap_social}, the positive and negative mean SHAP values for the social features are visualised.\\

\noindent\textbf{Hashtag count and posted day are good discriminators.} In Figure \ref{fig:mean_shap_social}, the base model CT consisting of content and temporal features indicate that \textit{hashtag count} and \textit{posted day} are good discriminators. The reason is two-fold: firstly, they have high positive and negative means (e.g. the bars are large) and secondly, the magnitude of the positive and negative mean is similar, meaning that features can affect a prediction in a positive and negative direction, equally. The feature \textit{users tagged} also has a high impact on the prediction, but the effect is mainly in a positive direction, since the positive mean is of larger magnitude than the negative mean and, consequently, it is not as good a discriminator as the two aforementioned. 
Regarding the size of the bars, similar trends from the top features in Figure \ref{fig:top_features} are observed in the figure.\\
\newline
\textbf{Language is important with visual features.} If we consider the first column in Figure \ref{fig:mean_shap_social}, only small changes are observed down the rows. The size of the bars is decreasing slightly as we add visual features, e.g \textit{word count} is larger in CT than YEPCT. Adding Y only seem to have very small effects on the bars and is not changing the relative distribution, whereas adding E and P give an increase in the positive mean of \textit{language}. In fact, all the features are smaller in YEPCT than in CT except \textit{language}, which is slightly higher. 
A similar trend is observed in the last two columns, where IIPA (I) is added to CT and ACT. I also affects the positive mean of \textit{language} in a positive direction, e.g. comparing CT with ICT. This is also seen for the other rows though the increase is smaller due to the increase from E and P. This indicates that \textit{language} is more important when visual semantics and I are added to the model. We hypothesise that the visual predictors of popularity vary across cultures.
\\
\newline
\textbf{The caption is less important with visual features.} If we compare the models in the first row with the models in the last row in Figure \ref{fig:mean_shap_social}, the attribution of the feature \textit{word count} has decreased. This indicates a connection between the visual features and the word count, which suggests that the visual information can partly substitute the information in the word count. Word count is the number of words in the caption, and thus, we observe how the caption is less important when visual features are present.\\ 
%
\newline
\textbf{Visual features have a small impact on social features.}
Overall, only small changes are observed across the models in Figure \ref{fig:mean_shap_social}, indicating that the visual features only slightly affect the impact of the social features on a prediction. If we compare the models in the first row and last row, the features \textit{language} has increased and \textit{word count} has decreased. If we compare ACT with YIEPACT, it is observed that the majority of the features have a smaller impact and \textit{word count} is very small but the author features \textit{followers} and \textit{followers per post} are unchanged, and the content feature \textit{language} is actually larger. This suggests that author features are important no matter the visual information, that \textit{language} might capture some sort of user segment, and that \textit{word count} and visual information are highly related.\\
\begin{table}[ht]
\scriptsize
\centering 
\caption{Ablation study with feature groups removed. Performance metrics are given by Spearman's rank correlation (SRC) and root mean square error (RMSE) together with the training and prediction time. All standard deviations with respect to RSME and SRC are below 0.002.}
\label{tab:ablation_study_small}
\begin{tabular}{c|rr|cc}
\hline
 &  \multicolumn{2}{c|}{Performance} & \multicolumn{2}{c}{Time}  \\ \hline
\textbf{Group removed} & SRC & RMSE & Train (s) & Pred. (ms) \\ \hline
Author          & 0.463 & 1.202 & 1075 & 186\\
EfficientNet     & 0.509 & 1.158 & 421 & 1055 \\
Places365        & 0.509 & 1.158 & 772 & 1111 \\
YOLOv3           & 0.510 & 1.157 & 1170 & 1051\\
IIPA             & 0.509 & 1.159 & 1105 & 1104\\ \hline         
\end{tabular}
\end{table}
\begin{figure}[t]
    \centering
    \includegraphics[width=0.98\columnwidth]{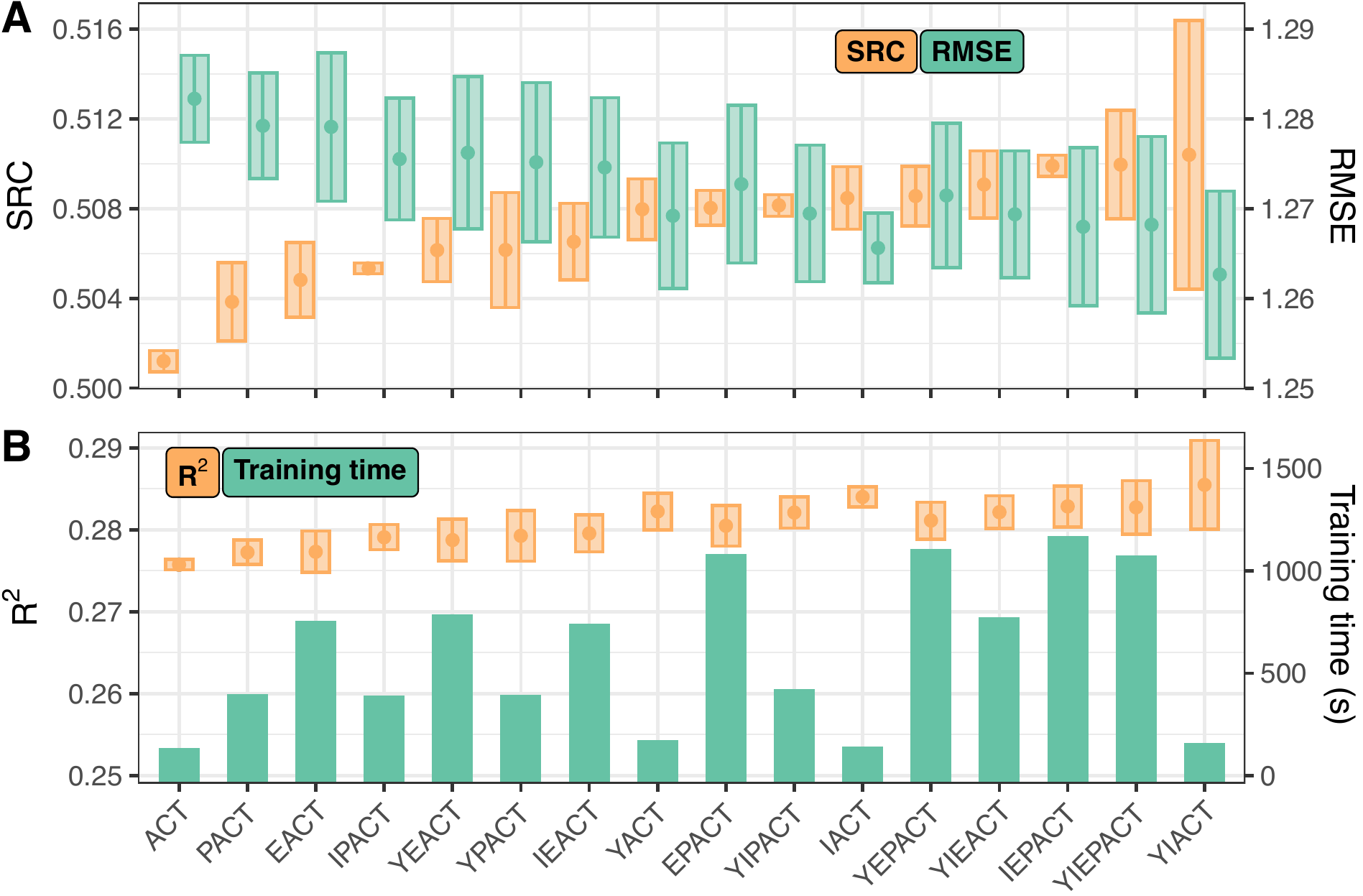}
    \caption{Performance for models getting an SRC higher than 0.5. The boxes shows $\pm2$ standard deviations. (A) Spearman's rank correlation (SRC) and Root Mean Square Error (RMSE). (B) $R^2$ and training time.}
    \label{fig:perf_best_models}
\end{figure}

\noindent The performance of the models is quantified using Spearman's rank correlation (SRC), Root Mean Square Error (RMSE), $R^2$, and training time. 
\begin{table}[ht!]
\scriptsize
\centering
\caption{Quantitative evaluation of all models given by Spearman's rank correlation (SRC), root mean square error (RMSE), the $R^2$, and the prediction time given in milliseconds. Abbr.: author (A), content (C), temporal (T), EfficientNet (E), Places365 (P), YOLO (Y), and IIPA (I).}
\label{tab:ablation_study_large}
    \begin{tabular}{r|c|c|c|c|c|c|r}
    \hline
                & \multicolumn{2}{c|}{\textbf{SRC}} & \multicolumn{2}{c|}{\textbf{RMSE}} & \multicolumn{2}{c|}{$\textbf{\textit{R}}^2$} & \textbf{Time} \\ \hline
         & $\mu$ & $\sigma$ & $\mu$ & $\sigma$ & $\mu$ & $\sigma$ & ms \\ \hline
        T & .261 & .001 & 1.306 & .001 & .086 & .001 & $<$1 \\ 
        C & .305 & .002 & 1.291 & .001 & .108 & .001 & $<$1 \\ 
        A & .349 & .002 & 1.266 & .001 & .141 & .001 & 935 \\ \hdashline
        CT & .417 & .001 & 1.231 & .001 & .188 & .000 & $<$1 \\ 
        AT & .425 & .001 & 1.219 & .002 & .204 & .001 & 936 \\ 
        AC & .426 & .000 & 1.216 & .001 & .207 & .000 & 936 \\ \hline
        \textbf{CT}  \\ \hline
        YCT & .433 & .000 & 1.222 & .001 & .200 & .000 & 71 \\ 
        ICT & .435 & .001 & 1.219 & .001 & .204 & .000 & 18 \\ 
        YICT & .444 & .001 & 1.214 & .001 & .211 & .001 & 88 \\ \hdashline
        PCT & .452 & .001 & 1.210 & .001 & .216 & .001 & 33 \\ 
        ECT & .455 & .000 & 1.208 & .001 & .219 & .001 & 89 \\ 
        YPCT & .456 & .000 & 1.207 & .002 & .220 & .001 & 103 \\ \hdashline
        IPCT & .456 & .000 & 1.206 & .001 & .221 & .001 & 50 \\ 
        YECT & .457 & .000 & 1.206 & .002 & .221 & .001 & 159 \\ 
        IECT & .458 & .001 & 1.205 & .001 & .222 & .000 & 106 \\  \hdashline
        YIPCT & .459 & .000 & 1.204 & .001 & .224 & .001 & 120 \\ 
        EPCT & .460 & .001 & 1.205 & .001 & .223 & .000 & 99 \\ 
        YIECT & .461 & .000 & 1.204 & .001 & .224 & .001 & 176 \\ \hdashline
        YEPCT & .461 & .000 & 1.204 & .002 & .224 & .001 & 169 \\ 
        IEPCT & .462 & .001 & 1.202 & .001 & .226 & .001 & 116 \\ 
        YIEPCT & .463 & .000 & 1.202 & .001 & .227 & .001 & 186 \\ \hline
        \textbf{ACT} \\ \hline
        ACT & .501 & .000 & 1.163 & .001 & .276 & .000 & 936 \\
        PACT & .504 & .001 & 1.162 & .001 & .277 & .001 & 968 \\
        EACT & .505 & .001 & 1.162 & .002 & .277 & .001 & 1024 \\ \hdashline
        IPACT & .505 & .000 & 1.160 & .001 & .279 & .001 & 985 \\
        YEACT & .506 & .001 & 1.160 & .002 & .279 & .001 & 1094 \\ 
        YPACT & .506 & .001 & 1.160 & .002 & .279 & .002 & 1038 \\ \hdashline
        IEACT & .507 & .001 & 1.160 & .002 & .280 & .001 & 1041 \\ 
        YACT & .508 & .001 & 1.158 & .002 & .282 & .001 & 1006 \\ 
        EPACT & .508 & .000 & 1.159 & .002 & .280 & .001 & 1034 \\ \hdashline
        YIPACT & .508 & .000 & 1.158 & .002 & .282 & .001 & 1055 \\ 
        IACT &   .508 & .001 & 1.156 & .001 & .284 & .001 & 954 \\
        YEPACT & .509 & .001 & 1.159 & .002 & .281 & .001 & 1104 \\ \hdashline
        YIEACT & .509 & .001 & 1.158 & .001 & .282 & .001 & 1111 \\ 
        IEPACT & \textbf{.510} & .000 & 1.157 & .002 & .283 & .001 & 1051 \\
        YIEPACT & \textbf{.510} & .001 & 1.157 & .002 & .283 & .002 & 1121 \\
        YIACT & \textbf{.510} & .003 & \textbf{1.155} & .002 & \textbf{.285} & .003 & 1023 \\ \hline
    \end{tabular}
\end{table}
In the top panel of Figure \ref{fig:perf_best_models}, the performance $\pm2$ standard deviations for the 16 best models are shown. As expected, the SRC and RMSE are inversely related. The standard deviations of performance between cross-validation folds form a conservative (too large) estimate of the standard error of the mean.
YIACT has the highest SRC, but also a high standard deviation, while the model IEPACT has similar performance but is more robust.
If we also include the $R^2$ and the training time from the bottom panel of Figure \ref{fig:perf_best_models}, we note that the models ACT, YACT, IACT, and YIACT are fast with training times below 200 seconds. All the other models have more than four times as many features, which is reflected in the increased training time. If $R^2$ is also taken into account, YIACT has the highest values but IACT has similar performance with much lower standard deviation. 
The model IACT has a low training time, a high $R^2$, and a high SRC with a small confidence interval. Hence, it is a good candidate for a strong, robust, and efficient baseline for Instagram popularity prediction. If we accept the somewhat larger training time (about 20 minutes), the model IEPACT is an excellent and robust candidate with a strong, consistent SRC performance across cross-validation folds.
For a real-time application, the prediction time is a central metric. The prediction time includes the feature extraction, and we assume that if you want to predict the popularity of a new post, you have the image, content, and temporal information at hand. The author features are crawled from WWW and the visual features are obtained via a propagation through the networks. In parallel, all LightGBM models run in less than one tenth of a millisecond. In Table \ref{tab:ablation_study_small} and Table \ref{tab:ablation_study_large}, the prediction time for a single evaluation of a post is seen. 
%
%
%
%
\section{\uppercase{Conclusions}}
In this paper, we revisit the problem of content popularity ranking on Instagram with a general population-based approach and no prior information about the content's authors.
We use a multi-modal approach to popularity prediction and focus on enhancing the visual modality's predictive power alongside the model's explainability, scalability, and robustness. We design a comprehensive ablation study including transfer learning to represent visual semantics with the explainable features concepts, scenes, and objects. The approach is strong, since we show robustness and consistency across models that take advantage of the synergy between the visual semantics. 
We show that the approach is explainable on both a high-level with feature groups and a low-level with individual features. We use Shapley analysis to quantify each feature's impact on the predictions. We calculate Shapley values for every prediction, before aggregating the explanations to provide novel attributions for all the visual semantics detected. In particular, we find that object detection works better with author features, and language is important with visual semantics. 
\newline\indent
Finally, we recommend two strong, explainable and scalable baselines which also inform a new lower limit in popularity ranking on Instagram, with population-based approach and without prior author information. 
We can lower bound the predictability as Spearman's rank correlation (SRC) $> 0.5$.
%
Based on the many combinations of multi-modal models, we make the following recommendations: If training time is of importance, we recommend the model (IACT) that combines author, content and temporal features with a single dimension measure of image popularity. This model trains in less than three minutes. If the focus is on robust performance and less on time to train, we recommend the model (IEPACT) that combines the social features with intrinsic image popularity and visual embeddings from EfficientNet and Places, which is about seven times slower in training. However, the latter model shows both strong and consistent SRC across cross-validation folds.
\newline\indent
Immediate avenues of future inquiry include experiments to explain how the impact of visual semantics varies across languages or investigating why object detection performs better with author information. Separately, it would be of high interest to apply the proposed visual feature extraction across population segments and social media platforms. Eventually, we hope to inspire further applications of explainable transfer learning to computational social science at scale.
\section*{\uppercase{Acknowledgments}}
This project is supported by the Innovation Fund Denmark, the Danish Center for Big Data Analytics driven Innovation (DABAI) and the Business Applications Group within Microsoft.
%
\bibliographystyle{apalike}
{\small\bibliography{bibliography}}

\end{document}